\pdfoutput=1
\documentclass[11pt,table]{article}

\usepackage[]{EMNLP2022}

\usepackage{times}
\usepackage{latexsym}

\usepackage[T1]{fontenc}
\usepackage[utf8]{inputenc}

\usepackage{microtype}

\usepackage{inconsolata}
\usepackage{xcolor}
\usepackage[normalem]{ulem}
\usepackage{graphicx}
\usepackage{multirow}
\usepackage{lscape}
\usepackage{booktabs}

\newcommand\correspondingauthor{\thanks{* Corresponding Author}}

\title{Knowledge Distillation Transfer Sets and their Impact on Downstream NLU Tasks}

\author{
  Charith Peris \correspondingauthor\\
  Amazon, Cambridge, USA \\
  \texttt{perisc@amazon.com} \\
  \And
  Lizhen Tan \\
  Amazon, Cambridge, USA \\
  \texttt{ltn@amazon.com} \\
  \And
  Thomas Gueudre \\
  Amazon, Turin, Italy \\
  \texttt{tgueudre@amazon.it} \\
  \AND
  Turan Gojayev \\
  Amazon, Berlin, Germany \\
  \texttt{tgojayev@amazon.de} \\
  \And
  Pan Wei \\
  Amazon, Cambridge, USA \\
  \texttt{panwei@amazon.com} \\
  \And
  Gokmen Oz\\
  Amazon, Cambridge, USA \\
  \texttt{ogokmen@amazon.com} \\
}

\begin{document}
\maketitle

\begin{abstract}
Teacher-student knowledge distillation is a popular technique for compressing today's prevailing large language models into manageable sizes that fit low-latency downstream applications. Both the teacher and the choice of transfer set used for distillation are crucial ingredients in creating a high quality student. Yet, the generic corpora used to pretrain the teacher and the corpora associated with the downstream target domain are often significantly different, which raises a natural question: {\it should the student be distilled over the generic corpora, so as to learn from high-quality teacher predictions, or over the downstream task corpora to align with finetuning?} Our study investigates this trade-off using Domain Classification (DC) and Intent Classification/Named Entity Recognition (ICNER) as downstream tasks. We distill several multilingual students from a larger multilingual LM with varying proportions of generic and task-specific datasets, and report their performance after finetuning on DC and ICNER. We observe significant improvements across tasks and test sets when only task-specific corpora is used. We also report on how the impact of adding task-specific data to the transfer set correlates with the similarity between generic and task-specific data. Our results clearly indicate that, while distillation from a generic LM benefits downstream tasks, students learn better using target domain data even if it comes at the price of noisier teacher predictions. In other words, target domain data still trumps teacher knowledge.

\end{abstract}

\section{Introduction}

In the recent past, large language models (LMs; BERT-Large, \citealp{devlin-etal-2019-bert}; GPT-2, \citealp{Radford2019LanguageMA}; T5, \citealp{JMLR:v21:20-074}) pretrained in a self-supervised manner on massive web corpora have consistently shown state-of-the-art performance for multiple natural language understanding (NLU) tasks. Therefore, it is no surprise that these models are of much interest for virtual assistants such as Amazon Alexa, Apple Siri, and Google Assistant. Some studies have shown that these large models trained on generic corpora seem to be more robust to data distributional shifts, relying less on domain-specific training data to perform well \citep{NEURIPS2020_1457c0d6}.

Since large models cannot be directly used for low-latency applications on devices with limited computing capacity, many techniques have been developed to compress them in size. Knowledge distillation (referred to simply as distillation hereafter; \citealp{hinton2015distilling}), has shown promising results, especially at the high compression rates typically required in NLU (\citealp{jiao-etal-2020-tinybert}, \citealp{soltan-etal-2021-limitations}). In this paradigm, lightweight models referred to as students, are trained to mimic the teacher predictions over a transfer set \citep{hinton2015distilling}. When the pretraining and task-specific corpora have significantly different distributions, as is often the case, the choice of data for the transfer set can be ambiguous. On the one hand, using pretraining corpora in the transfer set ensures high quality teacher predictions that are important for effective distillation. On the other, using the downstream corpora, although it might cause noisier teacher predictions, ensures the adaptation of the student to its final use case.

To investigate this trade-off, we present a set of experiments where we distill several multilingual students from a large multilingual teacher LM trained using a masked language modeling (MLM) objective. We perform the distillations using transfer sets that comprise of generic and task-specific data in varying proportions. The students are then finetuned and evaluated on two downstream NLU tasks of interest: a Domain Classification (DC) task and a joint Intent Classification/Named Entity Recognition (ICNER) task. For each input utterance DC predicts the relevant domain (Books, Music, Shopping, etc.), IC identifies the user's intent (find a book, play a song, buy an item, etc.) and NER extracts the entities in the utterance (dates, names, locations, etc.).

{\bf Our contributions}: (1) We confirm for our setup that model preparation via distillation from a larger LM is more beneficial for downstream task performance when compared to encoder training from scratch. (2) We show that the largest improvements are seen when using only the downstream task's unlabelled data during the distillation process. Even though teacher predictions are expected to be noisy over data that is different from pretraining corpora, our results clearly indicate that students learn best in this setting. (3) Because our ICNER corpora is divided per domain, we are also able to provide a finer-grained analysis of the impact of corpora similarity on downstream results. (4) Finally, we also confirm that further adaptation of the teacher to the target-domain data, results in improved student performance across tasks.

%The results we present might prove useful for teams at the onset of their distillation ventures, who wish to understand the effects of bolstering generic datasets with available target-domain data.

\section{Relevant Work}

Building models with inference speeds that are suitable for production systems is of utmost importance in the industrial setting. Therefore techniques for model compression (quantization \citealp{gong2014compressing}; pruning redundant connections \citealp{han2015learning}) have been active research topics, with distillation (\citealp{romero2015fitnets}, \citealp{hinton2015distilling}, \citealp{jiao-etal-2020-tinybert}) showing much promise for NLU models \citep{DistilBERT2019}. Distillation processes and their data have evolved over the past few years. In the teacher-student framework proposed by \citet{hinton2015distilling}, they recommend using the original pretraining set as the transfer set. \citet{jiao-etal-2020-tinybert} proposes a more complex two-stage process with generic and task-specific distillation phases, each with their own data sets, designed to augment the performance of the final model towards the task at hand. 

Our work is focused on exploring how varying proportions of generic and task-specific data within the transfer set of a single distillation process impacts downstream NLU performance. Since our scope does not include optimizing the distillation process itself, we use a cheaper alternative to \citet{jiao-etal-2020-tinybert}, via a single-stage distillation setup to conduct our exploration (see Section~\ref{app:student_setup} for details).

\citet{gururangan-etal-2020-dont} showed for the {\it pretraining} phase, that continued domain-adaptive and task-adaptive pretraining using the downstream task’s unlabeled data can improve performance. Our work presents similar results for the distillation phase.

\section{Data}
\subsection{Distillation data}
\label{dist_data}

For distillation, we created the transfer sets by mixing two types of data with different distributions:
\begin{itemize}
\item{{\bf Generic data:} This data set consisted of Wikipedia and Common Crawl processed by an in-house tokenizer.}
\item{{\bf Task-specific data:} This in-house data set comprised of de-identified utterances from a voice assistant across domains of interest. The text data collected here was the output of an Automatic Speech Recognition (ASR) model, which assigned a confidence score per utterance. In order to retain only the highest quality data, we filtered it by an ASR score threshold. The data was de-identified, prior to use. }
\end{itemize}

Our distilled students were trained as part of a larger program resulting in a collection of nine European and Indic languages being used for distillation. The language list and counts are shown in Table~\ref{tab:dist_data_counts}.

We built transfer sets that had three ratios of generic to task-specific data: (1) generic-only (baseline) (2) 7:3 generic to task-specific, to mimic the commonly encountered low task-specific data setting and (3) task-specific-only. To have a comparable distribution of data from each language, we created samples of equal size for each language using either generic only, task-specific only or combining both the generic and the task-specific data based on the targeted ratio. Upsampling is used when a source data set contains a number less than the required number. The 7:3 ratio consisted of Wikipedia, Common Crawl and task-specific data upsampled to counts of 35M, 35M and 30M respectively, for each language. For two languages Indian-English and Marathi, where some data constituents were unobtainable, available data was used in proportion (see Table~\ref{tab:dist_data_counts}). Once the data sets were created with the targeted mixing ratio, they were split into train and validation sets with a ratio of 0.995:0.005 and then used in the transfer sets. 

\subsection{Data for downstream tasks}
\label{eval_data}
We evaluated our multilingual distilled students in the context of two commonly utilized NLU tasks of interest, DC and ICNER. We limit the scope of our evaluation to just four languages German, French, Italian and Spanish. Our finetuning data consisted of 26 domains (see fractional utterance counts in Table~\ref{tab:fine-tune-data}) across each language, with each domain comprising a set of intents (similar to \citealt{Su2018ARS}). As with the task-specific data used in our transfer sets, this data has also been de-identified prior to use. 

It is important to note that, although collected over non-overlapping time intervals (and thus consisting of different absolute counts), the finetuning data was from the same distribution as the task-specific data described in Section~\ref{dist_data}. We sampled the finetuning data so as to have equal counts across each domain in all four languages (see Appendix~\ref{app:eval_set_creation} for the evaluation data set sampling strategy). We then combined all languages and split the data into proportions of 80:10:10 for train, validation and test, respectively. 

For the DC task, we classified the input utterances into one of the 26 domains. Therefore, the DC model is trained using the combined training data from the four languages across all domains and is tested on language-specific test data sets. For the joint ICNER task, we classified each utterance within a domain to its corresponding intent and also recognized its named-entities. For this task, we trained a model per domain, using the combined training data from the four languages for that domain. The model was evaluated using language-specific test data sets for that domain. We present results on two types of test sets. {\it test} comprises of the full test set obtained from the split above while {\it tail\_test} is the subset of data points within {\it test} that have a frequency of occurrence less than or equal to 3. The relative data proportions used can be found in the Appendix (Table~\ref{tab:fine-tune-data}).

\begin{figure*}
\centering
\includegraphics[width=1\textwidth]{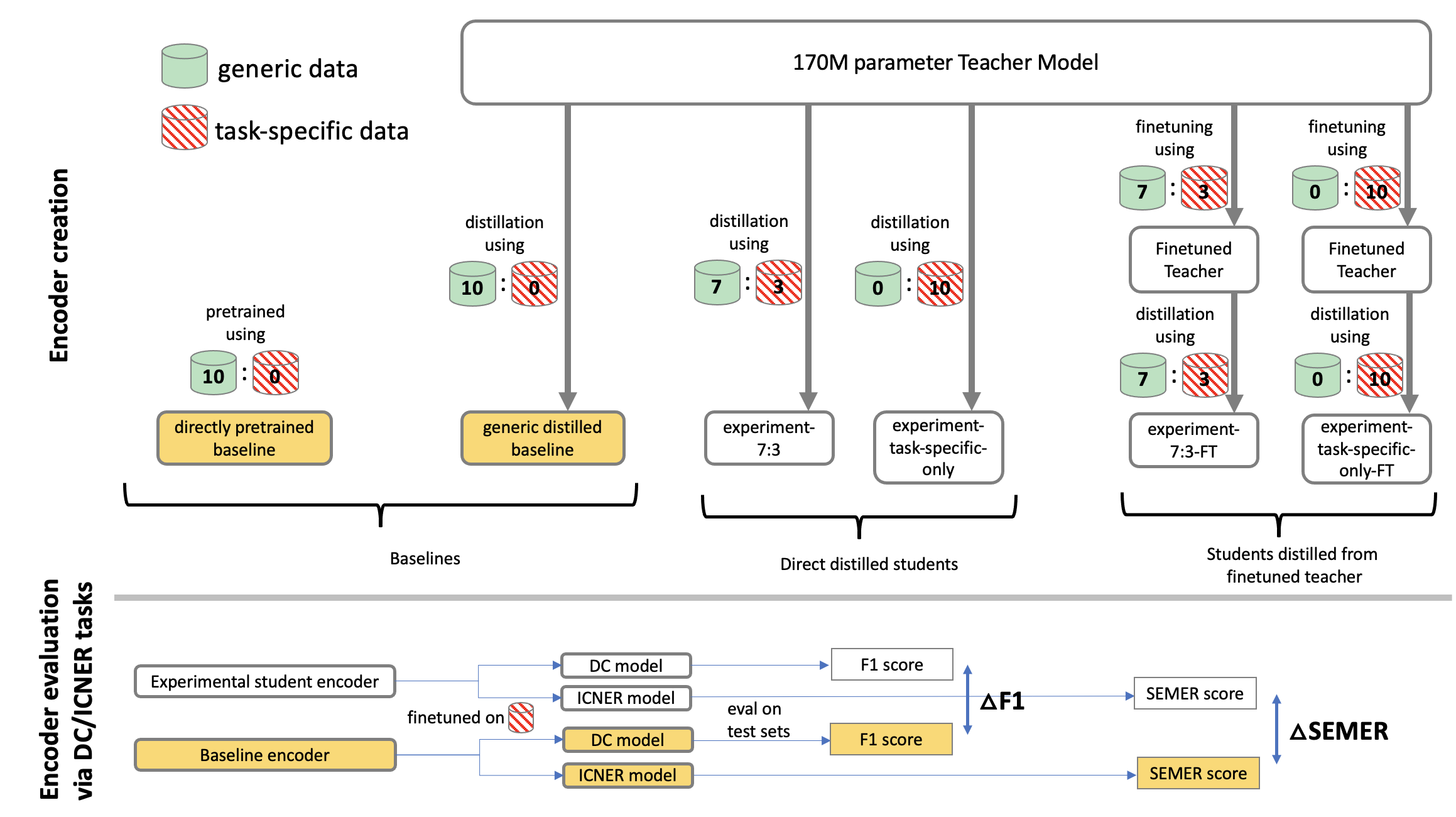}
\caption{A schematic of the models that we present in this paper and how they are evaluated.}
\label{fig:models}
\end{figure*}

\section{Models}
Figure~\ref{fig:models} shows a schematic of the models and experimental setup described in this section.

\subsection{Distilled students and baselines}
\label{sec:models}
We use a 170 million parameter teacher (170M-teacher) that was prepared using Wikipedia, Common Crawl and mC4 \citep{xue-etal-2021-mt5} data. See Appendix~\ref{app:teacher_model} for details on teacher preparation. From this teacher, we distilled a total of five students. We use our three transfer sets described in Section~\ref{dist_data}, i.e. (1) generic-only (2) 7:3 (generic:task-specific) and (3) task-specific-only, to distill the first three students. We refer to the student distilled using (1) as the generic-distilled baseline. The latter two are referred to as experiment-7:3 and experiment-task-specific-only; the naming aligned with the transfer set used. In addition to these, we create another two students where the teacher was finetuned using an MLM task before being used for distillation. In each case, the teacher was finetuned for 15625 steps using the same transfer set that was used for the subsequent distillation. We refer to these two students as experiment-7:3-FT and experiment-task-specific-only-FT. The teacher finetuning was run on a \texttt{p3.16X} instance with an average run time of approximately 45 hours. We collectively refer to all distilled students that are not a baseline as experimental students.

The architectures of our teacher and students are as follows. As in the paper by \citet{devlin-etal-2019-bert}, we denote the number of layers (i.e., Transformer blocks) as L, the hidden size as H, and the number of self-attention heads as A. 
\begin{itemize}
\item{170M-teacher: L=16, H=1024, A=16,  feed-forward/filter size=3072, total parameters=170M}
\item{Students: L=4, H=768, A=16,  feed-forward/filter size=1200, total parameters=17M}
\end{itemize}

For a description of the distillation setup, see Appendix~\ref{app:student_setup}. Distillation was run for 1 epoch with each student extracted at 78125 steps, which equates to approximately 80M data points seen. We ran distillation on a single \texttt{p3.16X} instance utilizing 8 GPUs with batch-size of 2 and gradient accumulation at every 64 steps. The average run time was approximately 195 hours. Note that each distillation run used only a sample of the full data set mentioned in Section~\ref{dist_data}, determined by the step count. However, since the data is sampled uniformly, the language ratios and the generic:task-specific data ratio stays consistent during training.

In addition to the distilled baseline, we also created another baseline ({\it without distillation}) that was directly pretrained using the generic-only data. The architecture and size of this baseline was identical to that of the distilled students and it is referred to, here onward, as the directly-pretrained baseline. We used this baseline to observe performance differences between models that use students distilled from the large teacher and those that use a directly pretrained encoder.

\subsection{DC and ICNER models} 

In order to evaluate the impact of the different transfer sets on our targeted downstream NLU tasks, we finetune the experimental students and baselines toward DC and ICNER tasks. Each DC model consisted of an encoder, embedding and positional embedding obtained from an experimental student or baseline combined with a decoder consisting of an MLP classifier for domain prediction with layer size 128, dropout set at 0.1 and ReLU activation. Each ICNER model consisted of the same encoder, embedding and positional embeddings used for the corresponding DC model with an MLP classifier output layer for the IC task with layer size 128, dropout set at 0.1 and ReLU activation and a CRF sequence-labeler output layer for the NER task with layer size 256, dropout set at 0.1 and GeLU activation. We trained each DC model for 1 epoch and each ICNER model for 4 epochs.

{\bf Evaluation}: The DC performance was evaluated using the F1 score while the ICNER performance was evaluated using the Semantic Error Rate (SemER; \citealp{Su2018ARS}, \citealp{varada-etal-2020-using}, \citealp{peris-etal-2020-using}). The definition of SemER is 

\begin{equation}
SemER = \frac{(D + I + S)}{(C + D + S)} 
\end{equation}

where D (deletion), I (insertion), S (substitution), C (correct slots). The Intent was treated as a slot in this metric, and the Intent error was considered as a substitution.

\begin{table*}[ht!]
\setlength{\tabcolsep}{4pt}
\centering
\small
\begin{tabular}{lccccccc}
\toprule
\textbf{Distilled encoder}  &  \textbf{Baseline}  &  \textbf{Test Set}  &  \textbf{German (\%)}  &  \textbf{French (\%)}  &  \textbf{Italian (\%)}  &  \textbf{Spanish (\%)}  \\
\midrule
experiment-7:3  &  generic distilled  &  test  &  $0.19\pm0.02$  &  $0.19\pm0.04$  &  $0.21\pm0.03$  &  $0.24\pm0.03$  \\
experiment-task-specific-only  &  generic distilled  &  test  &  $0.51\pm0.01$  &  $0.54\pm0.03$  &  $0.47\pm0.03$  &  $0.55\pm0.03$  \\
experiment-7:3-FT  &  generic distilled  &  test  &  $0.31\pm0.03$  &  $0.3\pm0.02$  &  $0.31\pm0.03$  &  $0.35\pm0.03$  \\
experiment-task-specific-only-FT  &  generic distilled  &  test  &  $\textbf{0.69}\pm\textbf{0.02}$  &  $\textbf{0.79}\pm\textbf{0.03}$  &  $\textbf{0.7}\pm\textbf{0.03}$  &  $\textbf{0.79}\pm\textbf{0.02}$  \\
\midrule
experiment-7:3  &  generic distilled  &  tail test  &  $0.34\pm0.05$  &  $0.31\pm0.09$  &  $0.42\pm0.05$  &  $0.42\pm0.05$  \\
experiment-task-specific-only  &  generic distilled  &  tail test  &  $1.0\pm0.03$  &  $1.05\pm0.04$  &  $1.02\pm0.06$  &  $1.07\pm0.06$  \\
experiment-7:3-FT  &  generic distilled  &  tail test  &  $0.56\pm0.06$  &  $0.61\pm0.04$  &  $0.65\pm0.06$  &  $0.6\pm0.07$  \\
experiment-task-specific-only-FT  &  generic distilled  &  tail test  &  $\textbf{1.38}\pm\textbf{0.05}$  &  $\textbf{1.51}\pm\textbf{0.06}$  &  $\textbf{1.48}\pm\textbf{0.06}$  &  $\textbf{1.51}\pm\textbf{0.05}$  \\
\bottomrule
\end{tabular}
\caption{Relative DC $\Delta$F1 ($\uparrow$), measured against the generic distilled baseline for each experimental student (positive is better). We run three iterations of each experimental student and show the percentage change of their means and its standard deviation.}
\label{tab:dc_results}
\end{table*}

\begin{table*}[ht!]
\setlength{\tabcolsep}{2.5pt}
\centering
\small
%\scriptsize
\begin{tabular}{lccccccc}
\toprule
\textbf{Distilled encoder}  &  \textbf{Baseline}  &  \textbf{Test Set}  &  \textbf{German (\%)}  &  \textbf{French (\%)}  &  \textbf{Italian (\%)}  &  \textbf{Spanish (\%)}  \\
\midrule
experiment-7:3  &  generic distilled  &  test  &  $-0.55\pm0.09$  &  $-0.31\pm0.07$  &  $-0.17\pm0.12$  &  $-0.17\pm0.09$  \\
experiment-task-specific-only  &  generic distilled  &  test  &  $-1.25\pm0.07$  &  $-0.81\pm0.07$  &  $-0.56\pm0.08$  &  $\textbf{-1.3}\pm\textbf{0.09}$  \\
experiment-7:3-FT  &  generic distilled  &  test  &  $-0.83\pm0.09$  &  $-0.49\pm0.13$  &  $-0.06\pm0.14$  &  $-0.58\pm0.09$  \\
experiment-task-specific-only-FT  &  generic distilled  &  test  &  $\textbf{-1.57}\pm\textbf{0.15}$  &  $\textbf{-1.18}\pm\textbf{0.07}$  &  $\textbf{-0.6}\pm\textbf{0.25}$  &  $-1.26\pm0.04$  \\
\midrule
experiment-7:3  &  generic distilled  &  tail test  &  $-0.49\pm0.07$  &  $-0.31\pm0.06$  &  $-0.16\pm0.09$  &  $-0.23\pm0.11$  \\
experiment-task-specific-only  &  generic distilled  &  tail test  &  $-1.19\pm0.05$  &  $-0.86\pm0.07$  &  $-0.67\pm0.08$  &  $\textbf{-1.44}\pm\textbf{0.09}$  \\
experiment-7:3-FT  &  generic distilled  &  tail test  &  $-0.83\pm0.09$  &  $-0.52\pm0.1$  &  $-0.13\pm0.09$  &  $-0.65\pm0.11$  \\
experiment-task-specific-only-FT  &  generic distilled  &  tail test  &  $\textbf{-1.53}\pm\textbf{0.12}$  &  $\textbf{-1.26}\pm\textbf{0.07}$  &  $\textbf{-0.79}\pm\textbf{0.14}$  &  $-1.32\pm0.06$  \\
\bottomrule
\end{tabular}
\caption{Relative ICNER $\Delta$SemER ($\downarrow$), measured against the generic distilled baseline for each experimental student (negative is better). As with DC, we run three iterations of the experimental students and show the percentage change of their means and its standard deviation. In calculating these percentage changes, we use the weighted average of the SemER for each domain in a given language, as the overall SemER in that language. }
\label{tab:ic_ner_results}
\end{table*}

\section{Experiments}
\subsection{Experimental results}
In this section, note that {\it model} refers a model that uses an experimental student or baseline encoder and has been finetuned towards a DC or ICNER task. Experimental models comprise of experimental student encoders and baseline models comprise of baseline encoders (see lower panel in Figure~\ref{fig:models}).

We used data across 26 domains to train and evaluate the DC and ICNER models (see Section~\ref{eval_data}). We compare the performance of each experimental model against the two baseline models (see Section~\ref{sec:models}). The improvements we quote in this section are $\Delta$F1 ($\uparrow$) (\textit{higher is better}) and $\Delta$SemER ($\downarrow$) (\textit{lower is better}; we use the weighted average of SemER across all domains) for DC and ICNER respectively, measured against the baseline models (Tables~\ref{tab:dc_results},~\ref{tab:ic_ner_results},~\ref{tab:dc_results_pt},~\ref{tab:ic_ner_results_pt}).

The results in Tables~\ref{tab:dc_results} and~\ref{tab:ic_ner_results} show that in general for both DC and ICNER tasks, all experimental students distilled with a mix of task-specific data (30\% or 100\%) perform significantly better than the generic distilled baseline. We further observe that models with encoders distilled with task-specific-only data yields the best overall performance which means that, in our setup, students learn better using target-domain data even if it comes at the price of noisier teacher predictions.

For all four languages across DC and for three out of four languages across ICNER, the best performances are observed with student models that were distilled from the {\it finetuned} teacher. This confirms that the additional step of finetuning the teacher and adapting it to the task-specific dataset, results in students that perform better on the intended downstream tasks.

We also note that across all task, language and test set combinations, the improvements seen against the directly pretrained baseline (see Tables~\ref{tab:dc_results_pt} and~\ref{tab:ic_ner_results_pt}) are larger than the improvements seen against the generic distilled baseline. For our setup, this shows that distilling from a large LM can benefit downstream tasks as opposed to using a similar-sized encoder pretrained from scratch; in other words our findings suggest that it is {\it better to distill} than to directly pretrain. However, we note that additional resources (in our case approximately 45 \texttt{p3.16X} hours) are required for this. 

The {\it tail\_test}, comprising of low frequency utterances within test, provides insights on the ability of the model to generalize to rarely seen utterances. For DC, we note that the improvements on {\it tail\_test} are significantly larger ($\sim$2X) than the improvements seen on the test set. This indicates that prediction on examples that appear infrequently in the task-specific data benefits more, from task-specific data being included in the distillation process. 

\begin{figure*}
\centering
\includegraphics[width=1\textwidth]{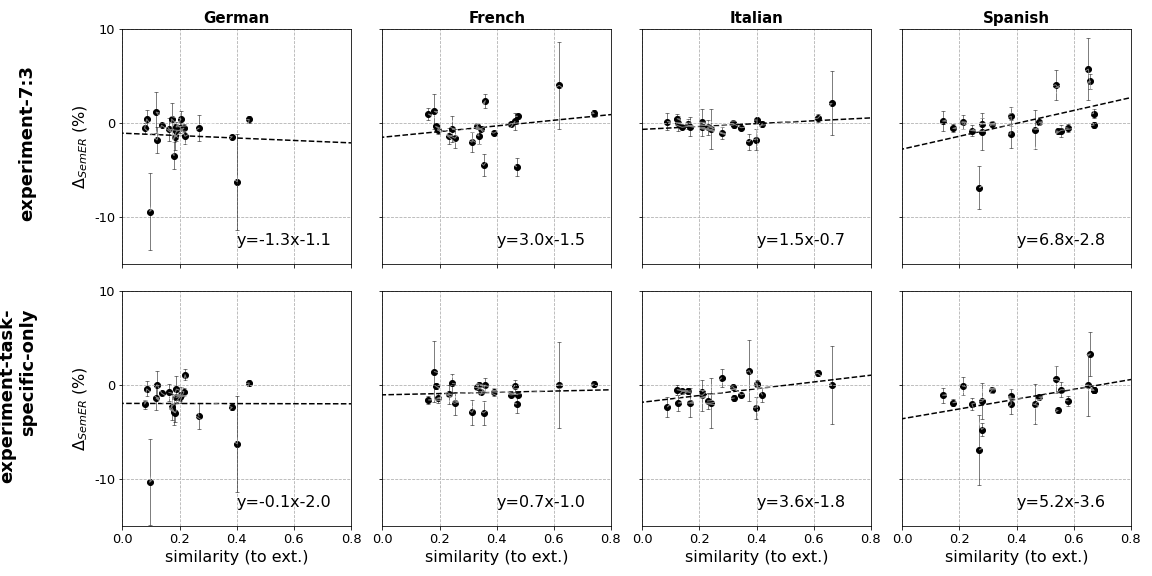}
\caption{Cosine similarity of {\it tf-idf} vectors vs. change in SemER for each domain for languages German, French, Italian and Spanish. We represent only domains with >1000 test utterances to avoid noise added by smaller domains which have higher variability.}
\label{fig:fig1}
\end{figure*}

\subsection{Dataset similarity and its correlation to SemER improvements for ICNER}

To further explore our conclusion that {\it students learn better using target-domain data} we explore how $\Delta$SemER for each domain, correlate to the similarity of the domain's data to the generic data. Note that, here, negative $\Delta$SemER represents improvements of the experimental students against the generic distilled baseline while the opposite is true for positive $\Delta$SemER. SemER results are from the {\it test} set. 

The hypothesis here is that the more distant a domain is from the generic data, the more value we should see in adding this domain's data to the distillation transfer set, even though teacher predictions might be noisy. We note here that we calculate cosine similarity on a very rudimentary corpus-level embedding (i.e. {\it tf-idf}) for measuring similarity, as explained below. We leave more sophisticated similarity measurements for later work.

To calculate similarity between domain-level and generic data, we use the following process. For each domain in each of the four languages, we sample up to 100K utterances. All available data is considered for domains with <100K utterances. We then sample 50K utterances each, from the Wikipedia and Common Crawl data sets of the corresponding language. We create a {\it tf-idf} vector for each sampled dataset and calculate their cosine similarity as a measure of dataset similarity. In order to account for any variability associated with the sampling, we repeat the process 3 times and obtain the mean similarity and the standard deviation per domain. We plot dataset similarity against $\Delta$SemER (a single point represents one domain and a panel represents a language as seen in Figure~\ref{fig:fig1}). We neglect domains with lower data and thus high variability and fit a line to show how $\Delta$SemER correlates to dataset similarity.

In Figure~\ref{fig:fig1}, we observe that a majority of cases (all except German) show a positive correlation. A positive correlation shows that domains that are {\it less} similar to Wikipedia/CommonCrawl have relatively larger improvement in SemER, when compared to domains that are more similar to Wikipedia/CommonCrawl. This suggests that the addition of task-specific data in the distillation transfer sets helps domains that are less similar to the generic data available for distillation, even though teacher predictions on them will be more noisy. 

It should be noted that the domains of the one exception, German, display low similarity values across the board unlike the other languages which show a wider spread (German has 65\% of domains < 0.2 whereas French, Italian and Spanish has 23\%, 31\% and 12\% < 0.2 respectively). The lack of domains with high similarity might explain the failure for a stable correlation to be observed in German.

\section{Conclusions}
We have explored how the use of transfer sets that comprise different ratios of generic to task-specific data, impacts downstream results. Encoders distilled from a large teacher perform better than ones trained from scratch, showing that it is {\it better to distill} than to directly pretrain, when possible. The largest benefits are shown when using the downstream task's unlabelled data to distill, a student despite noisier teacher predictions. We also find that domains with data that are dissimilar to the generic data show greater performance improvements against a generic baseline when using a student distilled using task-specific data. These improvements further confirm that distilling using target-domain data can be helpful for downstream performance. Finally, we show that if costs permit, teacher-adaptation to the target-domain data via finetuning can result in improved student performance across downstream tasks.

\section*{Acknowledgements}

We thank Karolina Owczarzak, Fabian Triefenbach and Rahul Gupta for their support of this work and Wael Hamza for helpful discussions on the topics covered.

\bibliography{anthology,custom}
\bibliographystyle{acl_natbib}

\appendix

\section{Appendix}
\label{sec:appendix}

\subsection{Data for finetuning DC and ICNER models}
\label{app:eval_set_creation}
For finetuning our distilled students for DC and ICNER, we use labelled datasets from four languages (German, French, Italian and Spanish) each consisting the same 26 domains (Table~\ref{tab:fine-tune-data}) and each domain supporting a set of intents (similar to \citealt{Su2018ARS}). In order to have equivalent utterance counts across domains for each language, we used a stratified sampling strategy as follows. First, we ranked each language per domain based on its utterance counts. In order to prevent heavy upsampling or downsampling in any single language when creating equivalently sampled domains, we picked the language that had the second highest utterance counts in most domains (in our case French). We sampled utterances from the domains of other languages to match the domain-level utterance frequency distribution of French (i.e. random sample utterances with replacement, from each domain in each language until that number matches the utterance count of the respective domain in French). We then combined all languages and split the data into proportions of 80:10:10 for train, validation and test, respectively.

\subsection{Teacher model}
\label{app:teacher_model}
The 170M-teacher used in this work was, itself, a student that was distilled from a larger model with 2 billion parameters (see Stage 1 pretraining section in \citet{FitzGerald2022AlexaTM} for details on creation and architecture). The 170M-teacher was distilled using a transfer set that comprised Wikipedia, Common Crawl and mC4 \citep{xue-etal-2021-mt5} data. Picking this intermediate-sized model helped us avoid potential performance degradation due to having too large a size gap between teacher and student \citep{mirzadeh2019improved}.

\subsection{Student setup}
\label{app:student_setup}
For our single-stage distillation setup, we skip the generic distillation phase done by \citet{jiao-etal-2020-tinybert} and use a non-finetuned teacher model to directly distill our students. In addition, as a sanity check, we also explore distillation from a finetuned teacher model to verify improved student performance across tasks. Similar to the hidden states based distillation followed in TinyBERT \citep{jiao-etal-2020-tinybert}, we mapped the student layers [0, 1, 2, 3] to learn from the teachers hidden layers [3, 7, 11, 15], respectively. We ignored attention based distillation \citep{jiao-etal-2020-tinybert} since we did not observe significant improvements by using it. We also penalized the soft cross-entropy loss between the student network’s logits against the teacher’s logits, to fit the students predictions to those of the teacher as in \citet{hinton2015distilling}. We use a MLM objective for the distillation process. In our loss, we weight the hidden layer matching, the logit matching and the MLM at a 1:2:1 ratio.

\setcounter{table}{0}
\renewcommand{\thetable}{A\arabic{table}}

\begin{table*}[ht!]
\setlength{\tabcolsep}{8pt}
\centering
\small
\begin{tabular}{lccc}
\toprule
       & \textbf{\begin{tabular}[c]{@{}c@{}}Common Crawl \\ (cc100)\end{tabular}} & \textbf{Wikipedia} & \textbf{Task Specific Data}  \\
\midrule
\textbf{German}       & 12,045,483                                                                                    & 2,731,840                               & 32,081,929              \\
\textbf{French}       & 13,323,804                                                                                    & 2,174,531                               & 21,278,820                \\
\textbf{Italian}      & 7,131,950                                                                                     & 1,278,255                               & 31,013,233                    \\
\textbf{Spanish}      & 11,690,123                                                                                    & 1,825,389                               & 22,054,722                \\ 
\textbf{English}      & 14,330,660                                                                                    & 6,360,372                               &  19,576,081                \\ 
\textbf{English (IN)} & -                                                                                              & -                                        & 27,406,082                             \\
\textbf{Hindi}        & 2,538,698                                                                                     & 94,891                                  & 21,315,004                     \\
\textbf{Tamil}        & 919,763                                                                                       & 66,190                                  &  -                                      \\
\textbf{Tamil (MT)}   & -                                                                                              &  -                                       & 18,414,285                             \\ 
\textbf{Telugu}       & 378,812                                                                                       & 77,179                                  &  -                                      \\
\textbf{Telugu (MT)}  & -                                                                                              & -                                        & 18,895,352                             \\
\textbf{Marathi}      & 263,189                                                                                       & 21,705                                  &  -                                      \\
\bottomrule
\end{tabular}
\caption{Raw data counts used for transfer set creation. Counts represent the number of sentences for generic data and the number of de-identified utterances for task-specific data. For task-specific data for Telugu and Tamil, machine-translated (MT) data from Indian English was used. Only task-specific data was used for Indian English because Wikipedia and Common Crawl data were not available. Only generic data was used for Marathi as the translation system used for this work did not support the language as yet.}
\label{tab:dist_data_counts}
\end{table*}

\begin{table*}[ht!]
\setlength{\tabcolsep}{4pt}
\centering
\small
\begin{tabular}{lccccccc}
\toprule
\textbf{Distilled encoder}  &  \textbf{Baseline}  &  \textbf{Test Set}  &  \textbf{German (\%)}  &  \textbf{French (\%)}  &  \textbf{Italian (\%)}  &  \textbf{Spanish (\%)}  \\
\midrule
experiment-7:3  &  directly pretrained  &  test  &  $0.31\pm0.02$  &  $0.19\pm0.04$  &  $0.32\pm0.01$  &  $0.28\pm0.03$  \\
experiment-task-specific-only  &  directly pretrained  &  test  &  $0.63\pm0.01$  &  $0.54\pm0.02$  &  $0.57\pm0.01$  &  $0.6\pm0.02$  \\
experiment-7:3-FT  &  directly pretrained  &  test  &  $0.42\pm0.03$  &  $0.3\pm0.02$  &  $0.41\pm0.01$  &  $0.4\pm0.02$  \\
experiment-task-specific-only-FT  &  directly pretrained  &  test  &  $\textbf{0.81}\pm\textbf{0.01}$  &  $\textbf{0.79}\pm\textbf{0.03}$  &  $\textbf{0.8}\pm\textbf{0.02}$  &  $\textbf{0.83}\pm\textbf{0.01}$  \\
\midrule
experiment-7:3  &  directly pretrained  &  tail test  &  $0.57\pm0.04$  &  $0.36\pm0.08$  &  $0.57\pm0.01$  &  $0.45\pm0.03$  \\
experiment-task-specific-only  &  directly pretrained  &  tail test  &  $1.23\pm0.02$  &  $1.1\pm0.02$  &  $1.17\pm0.03$  &  $1.09\pm0.04$  \\
experiment-7:3-FT  &  directly pretrained  &  tail test  &  $0.78\pm0.05$  &  $0.66\pm0.02$  &  $0.8\pm0.03$  &  $0.62\pm0.06$  \\
experiment-task-specific-only-FT  &  directly pretrained  &  tail test  &  $\textbf{1.61}\pm\textbf{0.04}$  &  $\textbf{1.56}\pm\textbf{0.05}$  &  $\textbf{1.63}\pm\textbf{0.03}$  &  $\textbf{1.54}\pm\textbf{0.04}$  \\
\bottomrule
\end{tabular}
\caption{Relative DC $\Delta$F1 ($\uparrow$) measured against the directly pretrained baseline for each experimental student (positive is better)}
\label{tab:dc_results_pt}
\end{table*}

\begin{table*}[ht!]
\setlength{\tabcolsep}{2.5pt}
\centering
\small
\begin{tabular}{lccccccc}
\toprule
\textbf{Distilled encoder}  &  \textbf{Baseline}  &  \textbf{Test Set}  &  \textbf{German (\%)}  &  \textbf{French (\%)}  &  \textbf{Italian (\%)}  &  \textbf{Spanish (\%)}  \\
\midrule

experiment-7:3  &  directly pretrained  &  test  &  $-1.73\pm0.06$  &  $-0.7\pm0.04$  &  $-2.3\pm0.11$  &  $-2.44\pm0.09$  \\
experiment-task-specific-only  &  directly pretrained  &  test  &  $-2.42\pm0.02$  &  $-1.19\pm0.04$  &  $-2.69\pm0.07$  &  $\textbf{-3.54}\pm\textbf{0.08}$  \\
experiment-7:3-FT  &  directly pretrained  &  test  &  $-2.01\pm0.06$  &  $-0.88\pm0.12$  &  $-2.2\pm0.13$  &  $-2.84\pm0.09$  \\
experiment-task-specific-only-FT  &  directly pretrained  &  test  &  $\textbf{-2.74}\pm\textbf{0.13}$  &  $\textbf{-1.57}\pm\textbf{0.05}$  &  $\textbf{-2.73}\pm\textbf{0.24}$  &  $-3.5\pm0.04$  \\
\midrule
experiment-7:3  &  directly pretrained  &  tail test  &  $-1.72\pm0.05$  &  $-0.7\pm0.05$  &  $-2.62\pm0.09$  &  $-2.62\pm0.09$  \\
experiment-task-specific-only  &  directly pretrained  &  tail test  &  $-2.41\pm0.02$  &  $-1.24\pm0.05$  &  $-3.11\pm0.07$  &  $\textbf{-3.8}\pm\textbf{0.07}$  \\
experiment-7:3-FT  &  directly pretrained  &  tail test  &  $-2.05\pm0.08$  &  $-0.9\pm0.09$  &  $-2.59\pm0.08$  &  $-3.03\pm0.09$  \\
experiment-task-specific-only-FT  &  directly pretrained  &  tail test  &  $\textbf{-2.74}\pm\textbf{0.11}$  &  $\textbf{-1.64}\pm\textbf{0.05}$  &  $\textbf{-3.23}\pm\textbf{0.13}$  &  $-3.68\pm0.03$  \\
\bottomrule
\end{tabular}
\caption{Relative ICNER $\Delta$SemER ($\downarrow$) measured against the directly pretrained baseline for each experimental student (negative is better). In calculating these percentage changes, we use the weighted average of the SemER for each domain in a given language, as the overall SemER in that language. }
\label{tab:ic_ner_results_pt}
\end{table*}

% \renewcommand{\thetable}{A}

% \usepackage{multirow}
\begin{landscape}
\begin{table}[]
\scriptsize
\caption{Fractions of finetuning data used per domain. Note that the faction in each cell represents the utterance count for that partition, for that domain, as a fraction of the total utterance count in that language. As mentioned in Section~\ref{eval_data}, these fractions are not based on the counts in Table~\ref{tab:dist_data_counts}}
\label{tab:fine-tune-data}
\begin{tabular}{|l|llll|llll|llll|llll|}
\hline
\multicolumn{1}{|c|}{\multirow{2}{*}{\textbf{}}} & \multicolumn{4}{c|}{\textbf{train}}                                                                                                          & \multicolumn{4}{c|}{\textbf{validation}}                                                                                                     & \multicolumn{4}{c|}{\textbf{test}}                                                                                                           & \multicolumn{4}{c|}{\textbf{tail\_test}}                                                                                                 \\ \cline{2-17} 
\multicolumn{1}{|c|}{}                           & \multicolumn{1}{l|}{\textbf{German}}   & \multicolumn{1}{l|}{\textbf{French}}   & \multicolumn{1}{l|}{\textbf{Italian}}  & \textbf{Spanish}  & \multicolumn{1}{l|}{\textbf{German}}   & \multicolumn{1}{l|}{\textbf{French}}   & \multicolumn{1}{l|}{\textbf{Italian}}  & \textbf{Spanish}  & \multicolumn{1}{l|}{\textbf{German}}   & \multicolumn{1}{l|}{\textbf{French}}   & \multicolumn{1}{l|}{\textbf{Italian}}  & \textbf{Spanish}  & \multicolumn{1}{l|}{\textbf{German}}  & \multicolumn{1}{l|}{\textbf{French}}  & \multicolumn{1}{l|}{\textbf{Italian}} & \textbf{Spanish} \\ \hline
\textbf{Domain 1}                                & \multicolumn{1}{l|}{0.014\%}           & \multicolumn{1}{l|}{0.014\%}           & \multicolumn{1}{l|}{0.014\%}           & 0.014\%           & \multicolumn{1}{l|}{0.002\%}           & \multicolumn{1}{l|}{0.002\%}           & \multicolumn{1}{l|}{0.002\%}           & 0.002\%           & \multicolumn{1}{l|}{0.002\%}           & \multicolumn{1}{l|}{0.002\%}           & \multicolumn{1}{l|}{0.002\%}           & 0.002\%           & \multicolumn{1}{l|}{0.002\%}          & \multicolumn{1}{l|}{0.002\%}          & \multicolumn{1}{l|}{0.002\%}          & 0.002\%          \\ \hline
\textbf{Domain 2}                                & \multicolumn{1}{l|}{0.634\%}           & \multicolumn{1}{l|}{0.634\%}           & \multicolumn{1}{l|}{0.630\%}           & 0.632\%           & \multicolumn{1}{l|}{0.078\%}           & \multicolumn{1}{l|}{0.079\%}           & \multicolumn{1}{l|}{0.078\%}           & 0.077\%           & \multicolumn{1}{l|}{0.081\%}           & \multicolumn{1}{l|}{0.080\%}           & \multicolumn{1}{l|}{0.080\%}           & 0.079\%           & \multicolumn{1}{l|}{0.059\%}          & \multicolumn{1}{l|}{0.063\%}          & \multicolumn{1}{l|}{0.058\%}          & 0.055\%          \\ \hline
\textbf{Domain 3}                                & \multicolumn{1}{l|}{0.663\%}           & \multicolumn{1}{l|}{0.664\%}           & \multicolumn{1}{l|}{0.662\%}           & 0.663\%           & \multicolumn{1}{l|}{0.082\%}           & \multicolumn{1}{l|}{0.082\%}           & \multicolumn{1}{l|}{0.082\%}           & 0.084\%           & \multicolumn{1}{l|}{0.083\%}           & \multicolumn{1}{l|}{0.083\%}           & \multicolumn{1}{l|}{0.082\%}           & 0.082\%           & \multicolumn{1}{l|}{0.059\%}          & \multicolumn{1}{l|}{0.064\%}          & \multicolumn{1}{l|}{0.062\%}          & 0.060\%          \\ \hline
\textbf{Domain 4}                                & \multicolumn{1}{l|}{0.072\%}           & \multicolumn{1}{l|}{0.072\%}           & \multicolumn{1}{l|}{0.072\%}           & 0.071\%           & \multicolumn{1}{l|}{0.009\%}           & \multicolumn{1}{l|}{0.009\%}           & \multicolumn{1}{l|}{0.009\%}           & 0.009\%           & \multicolumn{1}{l|}{0.009\%}           & \multicolumn{1}{l|}{0.009\%}           & \multicolumn{1}{l|}{0.009\%}           & 0.009\%           & \multicolumn{1}{l|}{0.007\%}          & \multicolumn{1}{l|}{0.008\%}          & \multicolumn{1}{l|}{0.007\%}          & 0.007\%          \\ \hline
\textbf{Domain 5}                                & \multicolumn{1}{l|}{6.024\%}           & \multicolumn{1}{l|}{6.037\%}           & \multicolumn{1}{l|}{6.024\%}           & 6.026\%           & \multicolumn{1}{l|}{0.760\%}           & \multicolumn{1}{l|}{0.754\%}           & \multicolumn{1}{l|}{0.757\%}           & 0.757\%           & \multicolumn{1}{l|}{0.755\%}           & \multicolumn{1}{l|}{0.751\%}           & \multicolumn{1}{l|}{0.756\%}           & 0.751\%           & \multicolumn{1}{l|}{0.346\%}          & \multicolumn{1}{l|}{0.343\%}          & \multicolumn{1}{l|}{0.355\%}          & 0.357\%          \\ \hline
\textbf{Domain 6}                                & \multicolumn{1}{l|}{0.524\%}           & \multicolumn{1}{l|}{0.525\%}           & \multicolumn{1}{l|}{0.525\%}           & 0.523\%           & \multicolumn{1}{l|}{0.067\%}           & \multicolumn{1}{l|}{0.065\%}           & \multicolumn{1}{l|}{0.065\%}           & 0.064\%           & \multicolumn{1}{l|}{0.067\%}           & \multicolumn{1}{l|}{0.066\%}           & \multicolumn{1}{l|}{0.065\%}           & 0.067\%           & \multicolumn{1}{l|}{0.028\%}          & \multicolumn{1}{l|}{0.033\%}          & \multicolumn{1}{l|}{0.036\%}          & 0.034\%          \\ \hline
\textbf{Domain 7}                                & \multicolumn{1}{l|}{0.369\%}           & \multicolumn{1}{l|}{0.369\%}           & \multicolumn{1}{l|}{0.368\%}           & 0.368\%           & \multicolumn{1}{l|}{0.045\%}           & \multicolumn{1}{l|}{0.047\%}           & \multicolumn{1}{l|}{0.046\%}           & 0.048\%           & \multicolumn{1}{l|}{0.047\%}           & \multicolumn{1}{l|}{0.047\%}           & \multicolumn{1}{l|}{0.046\%}           & 0.046\%           & \multicolumn{1}{l|}{0.031\%}          & \multicolumn{1}{l|}{0.036\%}          & \multicolumn{1}{l|}{0.027\%}          & 0.030\%          \\ \hline
\textbf{Domain 8}                                & \multicolumn{1}{l|}{1.043\%}           & \multicolumn{1}{l|}{1.042\%}           & \multicolumn{1}{l|}{1.041\%}           & 1.042\%           & \multicolumn{1}{l|}{0.131\%}           & \multicolumn{1}{l|}{0.131\%}           & \multicolumn{1}{l|}{0.132\%}           & 0.131\%           & \multicolumn{1}{l|}{0.133\%}           & \multicolumn{1}{l|}{0.128\%}           & \multicolumn{1}{l|}{0.132\%}           & 0.132\%           & \multicolumn{1}{l|}{0.077\%}          & \multicolumn{1}{l|}{0.060\%}          & \multicolumn{1}{l|}{0.066\%}          & 0.066\%          \\ \hline
\textbf{Domain 9}                                & \multicolumn{1}{l|}{18.474\%}          & \multicolumn{1}{l|}{18.462\%}          & \multicolumn{1}{l|}{18.464\%}          & 18.466\%          & \multicolumn{1}{l|}{2.307\%}           & \multicolumn{1}{l|}{2.300\%}           & \multicolumn{1}{l|}{2.299\%}           & 2.305\%           & \multicolumn{1}{l|}{2.305\%}           & \multicolumn{1}{l|}{2.310\%}           & \multicolumn{1}{l|}{2.312\%}           & 2.303\%           & \multicolumn{1}{l|}{0.202\%}          & \multicolumn{1}{l|}{0.263\%}          & \multicolumn{1}{l|}{0.342\%}          & 0.292\%          \\ \hline
\textbf{Domain 10}                               & \multicolumn{1}{l|}{0.005\%}           & \multicolumn{1}{l|}{0.005\%}           & \multicolumn{1}{l|}{0.005\%}           & 0.005\%           & \multicolumn{1}{l|}{0.001\%}           & \multicolumn{1}{l|}{0.000\%}           & \multicolumn{1}{l|}{0.001\%}           & 0.001\%           & \multicolumn{1}{l|}{0.001\%}           & \multicolumn{1}{l|}{0.001\%}           & \multicolumn{1}{l|}{0.001\%}           & 0.001\%           & \multicolumn{1}{l|}{0.001\%}          & \multicolumn{1}{l|}{0.001\%}          & \multicolumn{1}{l|}{0.001\%}          & 0.001\%          \\ \hline
\textbf{Domain 11}                               & \multicolumn{1}{l|}{0.462\%}           & \multicolumn{1}{l|}{0.465\%}           & \multicolumn{1}{l|}{0.464\%}           & 0.464\%           & \multicolumn{1}{l|}{0.057\%}           & \multicolumn{1}{l|}{0.058\%}           & \multicolumn{1}{l|}{0.057\%}           & 0.057\%           & \multicolumn{1}{l|}{0.057\%}           & \multicolumn{1}{l|}{0.059\%}           & \multicolumn{1}{l|}{0.059\%}           & 0.056\%           & \multicolumn{1}{l|}{0.029\%}          & \multicolumn{1}{l|}{0.034\%}          & \multicolumn{1}{l|}{0.021\%}          & 0.034\%          \\ \hline
\textbf{Domain 12}                               & \multicolumn{1}{l|}{9.737\%}           & \multicolumn{1}{l|}{9.720\%}           & \multicolumn{1}{l|}{9.735\%}           & 9.730\%           & \multicolumn{1}{l|}{1.218\%}           & \multicolumn{1}{l|}{1.217\%}           & \multicolumn{1}{l|}{1.215\%}           & 1.216\%           & \multicolumn{1}{l|}{1.216\%}           & \multicolumn{1}{l|}{1.214\%}           & \multicolumn{1}{l|}{1.209\%}           & 1.216\%           & \multicolumn{1}{l|}{0.401\%}          & \multicolumn{1}{l|}{0.373\%}          & \multicolumn{1}{l|}{0.352\%}          & 0.377\%          \\ \hline
\textbf{Domain 13}                               & \multicolumn{1}{l|}{5.915\%}           & \multicolumn{1}{l|}{5.916\%}           & \multicolumn{1}{l|}{5.923\%}           & 5.917\%           & \multicolumn{1}{l|}{0.741\%}           & \multicolumn{1}{l|}{0.740\%}           & \multicolumn{1}{l|}{0.737\%}           & 0.739\%           & \multicolumn{1}{l|}{0.736\%}           & \multicolumn{1}{l|}{0.735\%}           & \multicolumn{1}{l|}{0.733\%}           & 0.740\%           & \multicolumn{1}{l|}{0.511\%}          & \multicolumn{1}{l|}{0.497\%}          & \multicolumn{1}{l|}{0.478\%}          & 0.478\%          \\ \hline
\textbf{Domain 14}                               & \multicolumn{1}{l|}{1.454\%}           & \multicolumn{1}{l|}{1.456\%}           & \multicolumn{1}{l|}{1.457\%}           & 1.455\%           & \multicolumn{1}{l|}{0.183\%}           & \multicolumn{1}{l|}{0.184\%}           & \multicolumn{1}{l|}{0.183\%}           & 0.184\%           & \multicolumn{1}{l|}{0.181\%}           & \multicolumn{1}{l|}{0.182\%}           & \multicolumn{1}{l|}{0.183\%}           & 0.182\%           & \multicolumn{1}{l|}{0.162\%}          & \multicolumn{1}{l|}{0.171\%}          & \multicolumn{1}{l|}{0.172\%}          & 0.155\%          \\ \hline
\textbf{Domain 15}                               & \multicolumn{1}{l|}{13.658\%}          & \multicolumn{1}{l|}{13.667\%}          & \multicolumn{1}{l|}{13.666\%}          & 13.663\%          & \multicolumn{1}{l|}{1.703\%}           & \multicolumn{1}{l|}{1.717\%}           & \multicolumn{1}{l|}{1.705\%}           & 1.710\%           & \multicolumn{1}{l|}{1.709\%}           & \multicolumn{1}{l|}{1.710\%}           & \multicolumn{1}{l|}{1.714\%}           & 1.708\%           & \multicolumn{1}{l|}{0.998\%}          & \multicolumn{1}{l|}{1.101\%}          & \multicolumn{1}{l|}{1.100\%}          & 1.026\%          \\ \hline
\textbf{Domain 16}                               & \multicolumn{1}{l|}{3.470\%}           & \multicolumn{1}{l|}{3.471\%}           & \multicolumn{1}{l|}{3.472\%}           & 3.474\%           & \multicolumn{1}{l|}{0.435\%}           & \multicolumn{1}{l|}{0.434\%}           & \multicolumn{1}{l|}{0.437\%}           & 0.432\%           & \multicolumn{1}{l|}{0.435\%}           & \multicolumn{1}{l|}{0.435\%}           & \multicolumn{1}{l|}{0.435\%}           & 0.439\%           & \multicolumn{1}{l|}{0.209\%}          & \multicolumn{1}{l|}{0.245\%}          & \multicolumn{1}{l|}{0.231\%}          & 0.234\%          \\ \hline
\textbf{Domain 17}                               & \multicolumn{1}{l|}{0.028\%}           & \multicolumn{1}{l|}{0.029\%}           & \multicolumn{1}{l|}{0.029\%}           & 0.029\%           & \multicolumn{1}{l|}{0.004\%}           & \multicolumn{1}{l|}{0.004\%}           & \multicolumn{1}{l|}{0.004\%}           & 0.003\%           & \multicolumn{1}{l|}{0.004\%}           & \multicolumn{1}{l|}{0.003\%}           & \multicolumn{1}{l|}{0.004\%}           & 0.004\%           & \multicolumn{1}{l|}{0.003\%}          & \multicolumn{1}{l|}{0.003\%}          & \multicolumn{1}{l|}{0.003\%}          & 0.003\%          \\ \hline
\textbf{Domain 18}                               & \multicolumn{1}{l|}{0.655\%}           & \multicolumn{1}{l|}{0.656\%}           & \multicolumn{1}{l|}{0.657\%}           & 0.659\%           & \multicolumn{1}{l|}{0.082\%}           & \multicolumn{1}{l|}{0.083\%}           & \multicolumn{1}{l|}{0.083\%}           & 0.083\%           & \multicolumn{1}{l|}{0.082\%}           & \multicolumn{1}{l|}{0.082\%}           & \multicolumn{1}{l|}{0.084\%}           & 0.081\%           & \multicolumn{1}{l|}{0.026\%}          & \multicolumn{1}{l|}{0.029\%}          & \multicolumn{1}{l|}{0.030\%}          & 0.021\%          \\ \hline
\textbf{Domain 19}                               & \multicolumn{1}{l|}{0.428\%}           & \multicolumn{1}{l|}{0.429\%}           & \multicolumn{1}{l|}{0.428\%}           & 0.427\%           & \multicolumn{1}{l|}{0.053\%}           & \multicolumn{1}{l|}{0.053\%}           & \multicolumn{1}{l|}{0.053\%}           & 0.053\%           & \multicolumn{1}{l|}{0.055\%}           & \multicolumn{1}{l|}{0.053\%}           & \multicolumn{1}{l|}{0.053\%}           & 0.052\%           & \multicolumn{1}{l|}{0.046\%}          & \multicolumn{1}{l|}{0.046\%}          & \multicolumn{1}{l|}{0.046\%}          & 0.047\%          \\ \hline
\textbf{Domain 20}                               & \multicolumn{1}{l|}{10.209\%}          & \multicolumn{1}{l|}{10.210\%}          & \multicolumn{1}{l|}{10.211\%}          & 10.214\%          & \multicolumn{1}{l|}{1.271\%}           & \multicolumn{1}{l|}{1.276\%}           & \multicolumn{1}{l|}{1.283\%}           & 1.277\%           & \multicolumn{1}{l|}{1.273\%}           & \multicolumn{1}{l|}{1.277\%}           & \multicolumn{1}{l|}{1.272\%}           & 1.279\%           & \multicolumn{1}{l|}{0.678\%}          & \multicolumn{1}{l|}{0.642\%}          & \multicolumn{1}{l|}{0.706\%}          & 0.923\%          \\ \hline
\textbf{Domain 21}                               & \multicolumn{1}{l|}{0.046\%}           & \multicolumn{1}{l|}{0.045\%}           & \multicolumn{1}{l|}{0.045\%}           & 0.044\%           & \multicolumn{1}{l|}{0.006\%}           & \multicolumn{1}{l|}{0.006\%}           & \multicolumn{1}{l|}{0.005\%}           & 0.006\%           & \multicolumn{1}{l|}{0.006\%}           & \multicolumn{1}{l|}{0.006\%}           & \multicolumn{1}{l|}{0.006\%}           & 0.006\%           & \multicolumn{1}{l|}{0.003\%}          & \multicolumn{1}{l|}{0.003\%}          & \multicolumn{1}{l|}{0.003\%}          & 0.002\%          \\ \hline
\textbf{Domain 22}                               & \multicolumn{1}{l|}{1.247\%}           & \multicolumn{1}{l|}{1.253\%}           & \multicolumn{1}{l|}{1.250\%}           & 1.252\%           & \multicolumn{1}{l|}{0.159\%}           & \multicolumn{1}{l|}{0.155\%}           & \multicolumn{1}{l|}{0.159\%}           & 0.156\%           & \multicolumn{1}{l|}{0.156\%}           & \multicolumn{1}{l|}{0.160\%}           & \multicolumn{1}{l|}{0.157\%}           & 0.158\%           & \multicolumn{1}{l|}{0.128\%}          & \multicolumn{1}{l|}{0.129\%}          & \multicolumn{1}{l|}{0.107\%}          & 0.120\%          \\ \hline
\textbf{Domain 23}                               & \multicolumn{1}{l|}{0.223\%}           & \multicolumn{1}{l|}{0.223\%}           & \multicolumn{1}{l|}{0.222\%}           & 0.221\%           & \multicolumn{1}{l|}{0.028\%}           & \multicolumn{1}{l|}{0.029\%}           & \multicolumn{1}{l|}{0.029\%}           & 0.028\%           & \multicolumn{1}{l|}{0.028\%}           & \multicolumn{1}{l|}{0.028\%}           & \multicolumn{1}{l|}{0.029\%}           & 0.028\%           & \multicolumn{1}{l|}{0.028\%}          & \multicolumn{1}{l|}{0.026\%}          & \multicolumn{1}{l|}{0.026\%}          & 0.025\%          \\ \hline
\textbf{Domain 24}                               & \multicolumn{1}{l|}{2.986\%}           & \multicolumn{1}{l|}{2.981\%}           & \multicolumn{1}{l|}{2.978\%}           & 2.984\%           & \multicolumn{1}{l|}{0.372\%}           & \multicolumn{1}{l|}{0.369\%}           & \multicolumn{1}{l|}{0.372\%}           & 0.370\%           & \multicolumn{1}{l|}{0.373\%}           & \multicolumn{1}{l|}{0.373\%}           & \multicolumn{1}{l|}{0.372\%}           & 0.375\%           & \multicolumn{1}{l|}{0.265\%}          & \multicolumn{1}{l|}{0.289\%}          & \multicolumn{1}{l|}{0.289\%}          & 0.270\%          \\ \hline
\textbf{Domain 25}                               & \multicolumn{1}{l|}{1.617\%}           & \multicolumn{1}{l|}{1.614\%}           & \multicolumn{1}{l|}{1.617\%}           & 1.613\%           & \multicolumn{1}{l|}{0.202\%}           & \multicolumn{1}{l|}{0.203\%}           & \multicolumn{1}{l|}{0.204\%}           & 0.203\%           & \multicolumn{1}{l|}{0.202\%}           & \multicolumn{1}{l|}{0.202\%}           & \multicolumn{1}{l|}{0.201\%}           & 0.202\%           & \multicolumn{1}{l|}{0.083\%}          & \multicolumn{1}{l|}{0.091\%}          & \multicolumn{1}{l|}{0.098\%}          & 0.083\%          \\ \hline
\textbf{Domain 26}                               & \multicolumn{1}{l|}{0.042\%}           & \multicolumn{1}{l|}{0.042\%}           & \multicolumn{1}{l|}{0.042\%}           & 0.042\%           & \multicolumn{1}{l|}{0.006\%}           & \multicolumn{1}{l|}{0.005\%}           & \multicolumn{1}{l|}{0.005\%}           & 0.005\%           & \multicolumn{1}{l|}{0.005\%}           & \multicolumn{1}{l|}{0.005\%}           & \multicolumn{1}{l|}{0.006\%}           & 0.005\%           & \multicolumn{1}{l|}{0.004\%}          & \multicolumn{1}{l|}{0.003\%}          & \multicolumn{1}{l|}{0.004\%}          & 0.004\%          \\ \hline
\textbf{Total Fraction}                          & \multicolumn{1}{l|}{\textbf{80.000\%}} & \multicolumn{1}{l|}{\textbf{80.000\%}} & \multicolumn{1}{l|}{\textbf{80.000\%}} & \textbf{80.000\%} & \multicolumn{1}{l|}{\textbf{10.000\%}} & \multicolumn{1}{l|}{\textbf{10.000\%}} & \multicolumn{1}{l|}{\textbf{10.000\%}} & \textbf{10.000\%} & \multicolumn{1}{l|}{\textbf{10.000\%}} & \multicolumn{1}{l|}{\textbf{10.000\%}} & \multicolumn{1}{l|}{\textbf{10.000\%}} & \textbf{10.000\%} & \multicolumn{1}{l|}{\textbf{4.386\%}} & \multicolumn{1}{l|}{\textbf{4.554\%}} & \multicolumn{1}{l|}{\textbf{4.622\%}} & \textbf{4.708\%} \\ \hline
\end{tabular}
\end{table}
\end{landscape}

\end{document}